\documentclass[11pt]{article}

\usepackage[final]{acl}

\usepackage{times}
\usepackage{latexsym}
\usepackage{enumitem}
\usepackage{colortbl}
\usepackage{booktabs}
\usepackage{multirow}
\usepackage{bm}
\usepackage{adjustbox}
\usepackage{subcaption}
\usepackage{amsmath}
\usepackage{algorithm}
\usepackage{algpseudocode} 
\usepackage{amssymb}
\usepackage{tabularx}  

\usepackage{listings}

\lstdefinestyle{prompt}{
    basicstyle=\ttfamily\scriptsize,
    breaklines=true,
    breakatwhitespace=true,
    columns=fullflexible,
    keepspaces=true,
    showstringspaces=false,
    frame=single,
    framerule=0.3pt,
    xleftmargin=0.5em,
    xrightmargin=0.5em,
    aboveskip=0.6em,
    belowskip=0.6em
}

\usepackage{xcolor}
\usepackage{tikz}
\usepackage{pgfplots}
\pgfplotsset{compat=1.18}

\usepackage[T1]{fontenc}
\usepackage[utf8]{inputenc}

\usepackage{microtype}

\usepackage{inconsolata}

\usepackage{graphicx}
\usepackage{xspace}
\newcommand{\method}{MERIT\xspace}

\title{Causal Episodic Memory for Feedback-Driven Agent Repair}

\author{
    \textbf{Khang Nhat Hoang Vo}\textsuperscript{1} \quad
  \textbf{Tam Minh Chu}\textsuperscript{2} \quad
  \textbf{Anh Trac Duc Dinh}\textsuperscript{2} \\
  \textbf{Thuyen Vinh Ha Bui}\textsuperscript{2} \quad
  \textbf{Tho T. Quan}\textsuperscript{2}
\\
    \textsuperscript{1}Mohamed bin Zayed University of Artificial Intelligence, Abu Dhabi, United Arab Emirates 
    \\
  \textsuperscript{2}Faculty of Computer Science and Engineering, Ho Chi Minh City University of Technology \\ (HCMUT), VNU-HCM, Ho Chi Minh City, Vietnam
\\
  \small{\textbf{Correspondence:} \href{mailto:Khang.Vo@mbzuai.ac.ae}{Khang.Vo@mbzuai.ac.ae}, \href{mailto:qttho@hcmut.edu.vn}{qttho@hcmut.edu.vn}}
}

\begin{document}
\maketitle
\begin{abstract}
LLM agents that repair failures often discard successful corrections,
forcing later episodes to rediscover similar solutions. We study whether
finalized repair outcomes can improve subsequent Text-to-SQL episodes
without parameter updates. We introduce \method{}, a training-free agent
that maintains an online dual-polarity memory of oracle-verified
corrections and observed unsuccessful directions. Under oracle-assisted
benchmark feedback, only memories from earlier finalized episodes are
eligible for retrieval. A deterministic classifier assigns a coarse
failure type, which conditions a hybrid lexical-dense retriever before
the frozen model generates each revision. Using Qwen2.5-7B-Instruct with identical initial predictions and repair
budgets, \method{} improves execution accuracy\footnote{Throughout, ``execution accuracy'' refers to denotation-match accuracy under each benchmark's official evaluator; executable-but-incorrect SQL does not count as correct.} over stateless iterative
repair from \(66.34\%\) to \(69.79\%\) on Spider and from \(47.35\%\) to
\(48.44\%\) on BIRD. Paired analyses provide clear evidence for the
Spider gain but weaker evidence on BIRD. \method{} is not reliably
separated from untyped dynamic retrieval on either benchmark, while
Reflexion-style memory reaches \(51.24\%\) on BIRD at substantially
higher inference cost. Ablations show that negative memory contributes
modestly, the value of type conditioning and lexical-dense ranking is
dataset dependent, and schema-local experience provides the most
consistent benefit. These results clarify when causal cross-query memory
improves repair and when broader memory representations remain
preferable. Our implementation is available here: \url{https://github.com/nhatkhangcs/merit}
\end{abstract}
\section{Introduction}

Dependable language agents must do more than produce a correct action
once: they must respond to environmental feedback and retain what they
learn from it. ReAct established the interleaving of reasoning, acting,
and environment observations, while Self-Refine, CRITIC, and
Self-Debugging showed that iterative feedback, external tools, and
execution signals can improve an agent's current output
\cite{yao2023react,madaan2023selfrefine,gou2024critic,
chen2024teaching}. Such grounding is important because intrinsic
self-correction without external evidence is often unreliable and can
even degrade reasoning performance~\cite{huang2024cannot}. In
Text-to-SQL, decomposition and execution-guided refinement have likewise
improved generation by diagnosing or revising a query within its current
episode~\cite{pourreza2023dinsql,dai2026reexsql}. These methods,
however, largely treat each new query as a fresh problem. Once a useful
correction has been discovered, it is typically unavailable when a
similar failure appears in a later query. This limits an agent's ability
to improve through continued interaction, even when failures such as
missing joins, invalid schema references, and incorrect aggregations
recur across tasks.

Cross-episode memory offers a mechanism for learning from such
experience without modifying model parameters. Reflexion retains verbal
reflections from earlier trials, and ExpeL extracts reusable natural
language knowledge from collections of agent trajectories
\cite{shinn2023reflexion,zhao2024expel}. In Text-to-SQL, retrieval-based
methods such as DAIL-SQL and ACT-SQL instead select question-SQL
demonstrations to improve initial generation
\cite{gao2024dailsql,zhang2023actsql}. These approaches demonstrate the value of recalling prior experience,
but they do not organize repair memories jointly by the failure that
produced them and the outcome of the attempted correction. Semantic
similarity alone may therefore retrieve irrelevant episodes or conflate
unsuccessful attempts with verified corrections. We study whether structured episodic memory can improve cross-query
repair performance.

We introduce \textbf{MERIT}
(\textbf{M}emory-Augmented \textbf{E}rror-Typed \textbf{R}etrieval for
\textbf{I}terative \textbf{T}ext-to-SQL repair), a training-free agent
that accumulates structured repair experience online. MERIT stores
oracle-verified corrections as positive guidance and observed
unsuccessful directions as negative guidance. A deterministic classifier
assigns a coarse failure type, which conditions a hybrid
lexical--dense retriever before the frozen model generates each repair.
Only memories from finalized earlier episodes are available to the
current query, enabling improvement through cross-query experience
without parameter updates. We study this process under oracle-assisted benchmark feedback. Here,
\emph{causal} denotes temporal memory availability, not the absence of
correctness supervision. We make three contributions:

\begin{itemize}[noitemsep,topsep=0pt]
    \item We introduce MERIT and formulate Text-to-SQL correction as
    \emph{causal episodic memory for feedback-driven agent repair}, in
    which only outcomes from finalized earlier episodes may guide the
    current query.

\item We propose an error-typed hybrid retrieval design that organizes repair experience by outcome and failure type and ranks precedents using lexical and dense signals, and we empirically characterize when this organization helps relative to untyped retrieval and simpler memory baselines.

\item We provide a controlled evaluation over three stream orders, comparing accuracy, repair behavior, and inference cost across MERIT and alternative cross-query memory strategies, including cases where MERIT's structured organization is not reliably distinguished from untyped dynamic retrieval.
\end{itemize}

\section{Related Work}

\paragraph{LLM-based Text-to-SQL.}
Recent Text-to-SQL systems improve semantic parsing through structured
reasoning, candidate generation, and in-context example selection.
DIN-SQL decomposes complex questions into schema linking,
classification, and staged SQL generation~\cite{pourreza2023dinsql};
CHASE-SQL explores multiple reasoning paths and selects candidates using
execution-guided signals~\cite{pourreza2025chasesql}; and SQL-PaLM studies
instruction tuning and prompting at larger model scales
\cite{sun2024sqlpalm}. Retrieval-based prompting provides another source
of improvement: DAIL-SQL selects demonstrations using question and SQL
similarity, while ACT-SQL retrieves examples augmented with
automatically generated reasoning traces
\cite{gao2024dailsql,zhang2023actsql}. These methods primarily target
the quality of the initial prediction or select demonstrations from a
pre-constructed pool. MERIT instead begins from a failed prediction and
studies whether the correction discovered for one query can improve the
repair of later queries.

\paragraph{Execution-guided correction.}
Environment feedback provides an external signal for revising generated
programs and structured outputs. Self-Debugging uses execution results
to diagnose and revise code, Self-Refine iteratively improves outputs
through model-generated feedback, and CRITIC grounds correction in
tool-mediated critique
\cite{chen2024teaching,madaan2023selfrefine,gou2024critic}. Related work
has applied execution-guided diagnosis and repair directly to SQL
\cite{shen2026understanding,chau2025making,gong2025sqlens}. More specialized
Text-to-SQL agents introduce self-correction guidelines, multi-agent
refinement, search over candidate rewrites, or difficulty-aware
question rewriting
\cite{askari2025magic,deng2025reforce,
lyu2025sqlo1,mao2024dartsql}.
These approaches strengthen the correction process within a query.
MERIT addresses a complementary question: how that experience should be
retained and reused after the current query ends.

\paragraph{Experience retrieval and agent memory.}
Several agent architectures use memory to transfer information across
episodes. Reflexion converts feedback into verbal reflections that guide
later trials~\cite{shinn2023reflexion}, while ExpeL extracts reusable
knowledge from collections of successful and failed trajectories
\cite{zhao2024expel}. Retrieval-augmented repair
methods similarly expand an experience pool and retrieve related
examples for subsequent problems~\cite{zhao2025recode}. Closer to our setting, SQLFixAgent retrieves similar past repairs and reflects on failure memory to guide Text-to-SQL correction, while Memo-SQL reuses retrieved historical error-fix experience for training-free NL2SQL self-correction~\cite{10.1609/aaai.v39i1.31979, yang2026memosqlstructureddecompositionexperiencedriven}. Concurrent work, MIRA, similarly verifies repair evidence before it is stored in memory~\cite{liu2026miraevidenceverifiedrepairmemory}. MERIT differs from these systems through its online finalized-episode eligibility rule, explicit temporal information boundary, oracle-established memory polarity, frozen generator, and matched-stream evaluation protocol. 
General-purpose memory systems address longer-running interaction and
recall: MemGPT manages information across working and archival memory,
Generative Agents maintain experience streams and higher-level
reflections, and MemoryBank stores long-term conversational experience
\cite{packer2023memgpt,park2023generative,
zhong2024memorybank}. Together, these works establish that agents
can benefit from experience beyond the immediate context, but they do
not prescribe how execution-derived repair evidence should be
represented.

\section{Methodology}
\label{sec:method}

\begin{figure*}[t]
\centering
\includegraphics[width=\linewidth]{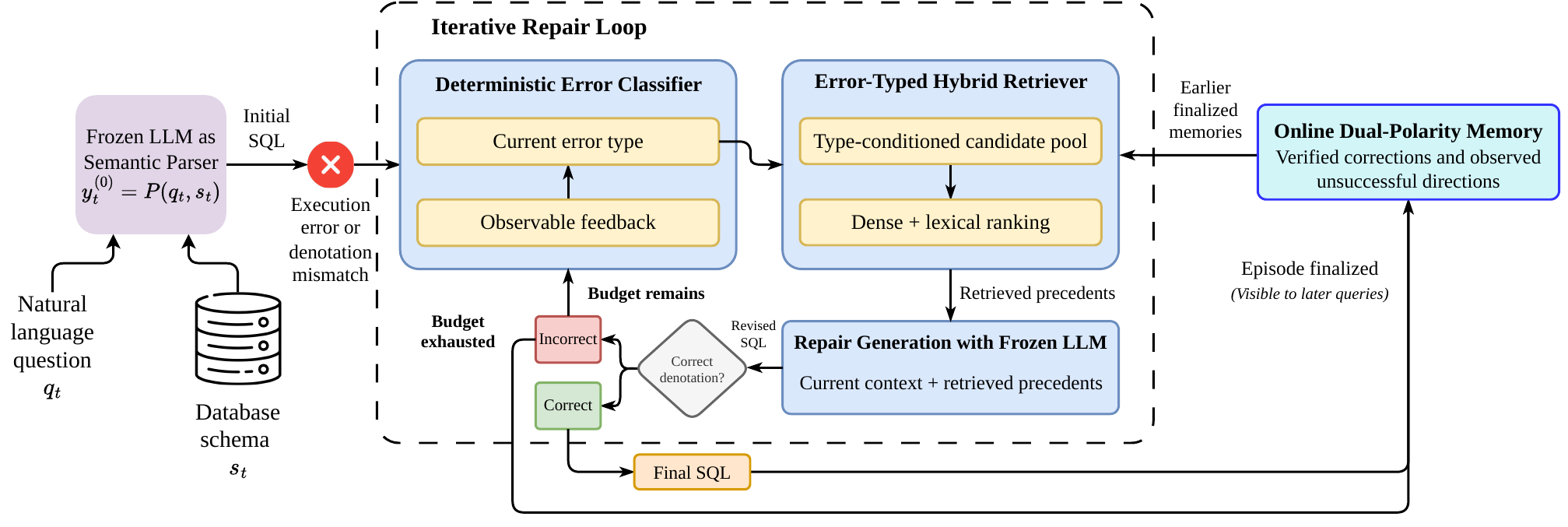}
\caption{
Overview of MERIT's causal online repair process. A frozen LLM produces
a shared initial SQL prediction. After an unsuccessful attempt, MERIT
classifies the observed failure, retrieves experience from earlier
finalized episodes, and generates one revision. A benchmark correctness
oracle controls termination and memory polarity without exposing the
reference SQL or result rows to the model. Current-query attempts remain
local, and each finalized episode contributes at most one memory entry
that becomes visible only to later queries.
}
\label{fig:overview}
\end{figure*}

\subsection{Causal Online Repair Formulation}
\label{subsec:problem_formulation}

Figure~\ref{fig:overview} presents MERIT as an online
correction--accumulation process. At stream position \(t\), the agent
receives a question \(q_t\), schema \(S_t\), optional dataset evidence
\(a_t\), executable database \(D_t\), and shared initial SQL prediction
\(y_t^{(0)}\). The prediction is generated once with greedy decoding and
reused by every repair method, isolating the effect of the repair
strategy.

\paragraph{Observable feedback and benchmark correctness.}
Our main experiments use the \texttt{denotation\_confirmed} regime.
After attempt \(k\), the benchmark evaluator returns
\(V(D_t,y_t^{(k)})=(z_t^{(k)},\epsilon_t^{(k)})\), where
\(z_t^{(k)}\) is \textsc{Correct}, \textsc{Denotation Mismatch},
\textsc{Execution Error}, or \textsc{Timeout}, and
\(\epsilon_t^{(k)}\) contains the latest observable Database Management System (DBMS) error when one
exists. Spider uses the pinned test-suite execution protocol, whereas
BIRD compares predicted and reference result rows under the pinned
timeout protocol.

The reference SQL and reference rows remain isolated inside the
evaluator and are not supplied to generation, classification, or
retrieval. The model receives only the attempt status, available DBMS
error, and predicted failure type. Nevertheless, the oracle decision
controls episode termination and memory polarity.

Each initially incorrect query defines one episode. After \(k\) repair
generations, its local history is
\(H_t^{(k)}=(y_t^{(0)},\ldots,y_t^{(k)})\), while
\(\mathcal{M}_{<t}\) contains entries produced by finalized episodes at
earlier stream positions. Given a repair budget \(K=7\), the objective
is to find some \(y_t^{(k)}\), \(k\leq K\), for which
\(z_t^{(k)}=\textsc{Correct}\).

Current-query attempts remain in \(H_t^{(k)}\) and cannot enter
cross-query retrieval. A memory entry \(x\) is eligible only if
\(\operatorname{sourcePos}(x)<t\) and
\(\operatorname{sourceQuery}(x)\neq q_t\), and global memory is updated
only after episode termination. Thus, \emph{causal} refers to temporal
memory availability: the current query cannot access its unfinished
trajectory or future-query experience. We enforce this boundary through four checks applied before every retrieval call: (i) a memory entry is inserted only after its source episode is finalized; (ii) entries produced by the current query's own attempts are excluded regardless of timestamp; (iii) candidate pools are filtered by $sourcePos(x) < t$; and (iv) reference SQL, reference rows, and corrected queries are never included in serialized memory text or prompts.

\subsection{MERIT: Correction and Experience Accumulation}
\label{subsec:merit}

MERIT combines a deterministic failure classifier, an online
dual-polarity memory, and a type-conditioned hybrid retriever. For each unsuccessful candidate, MERIT classifies the current failure,
retrieves causally available experience, and prompts the frozen model
to generate one revision. The episode continues until oracle-confirmed
correctness or budget exhaustion.

\paragraph{Initialization and failure diagnosis.}
If \(y_t^{(0)}\) is oracle-confirmed correct, the query is finalized
without creating repair memory because no corrective transition was
observed. Otherwise, MERIT initializes
\(H_t^{(0)}=[y_t^{(0)}]\) and enters the repair loop.

At step \(k\), the classifier maps the current unsuccessful query
\(y_t^{(k-1)}\) to a failure type \(\tau_t^{(k-1)}\). Its output
categories are \textsc{Result Mismatch}, \textsc{Schema Linking},
\textsc{Aggregation}, \textsc{Filter/Value}, \textsc{Execution},
\textsc{Syntax}, and \textsc{Unknown}. The classifier is deterministic
and uses only execution status and observable DBMS text. Ordered rules
identify schema-reference errors, parse failures, aggregate misuse, type
mismatches, execution failures, and timeouts; the first matching rule
determines the label. Other nonempty DBMS errors map to
\textsc{Unknown}, while executable but oracle-incorrect SQL maps to
\textsc{Result Mismatch}. The classifier does not inspect the SQL AST or
infer semantic errors such as incorrect joins or ordering. The label is
recomputed after each unsuccessful revision and serves only as a
retrieval prior. The complete rules appear in
Appendix~\ref{app:classifier-rules}.

\paragraph{Online dual-polarity memory.}
Memory is partitioned into positive and negative pools,
\(\mathcal{M}_{<t}=\mathcal{M}_{<t}^{+}\cup\mathcal{M}_{<t}^{-}\).
An entry is represented as
\(x=\langle c,\tau,\Delta,o,\rho\rangle\), where \(c\) stores the source
question, schema, unsuccessful SQL, status, and observable feedback;
\(\tau\) is the source failure type; \(\Delta\) describes the observed
SQL transformation; \(o\in\{+,-\}\) is the polarity; and \(\rho\)
records provenance.

A positive entry records an oracle-confirmed failed-to-correct
transition \(y^{-}\rightarrow y^{+}\). A negative entry records the
final direction attempted in an unresolved episode, indicating only
that it failed in its source context without asserting an unobserved
cause or universal invalidity.

\paragraph{Causal, type-conditioned retrieval.}
Candidate construction is performed independently for polarity
\(p\in\{+,-\}\). Let
\(\mathcal{L}_{t,p}\subseteq\mathcal{M}_{<t}^{p}\) be the legal
same-polarity pool after applying the temporal, query, and database
constraints. Its same-type subset is
\(\mathcal{T}_{t,p}=\{x\in\mathcal{L}_{t,p}:
\tau_x=\tau_t^{(k-1)}\}\). Full MERIT selects
\begin{equation}
\mathcal{C}_{t,p}
=
\begin{cases}
\mathcal{T}_{t,p},
& |\mathcal{T}_{t,p}|\geq 3,\\
\mathcal{L}_{t,p},
& \text{otherwise}.
\end{cases}
\label{eq:hard-type-filter}
\end{equation}

Thus, retrieval remains within the predicted type when at least three
same-type entries exist in that polarity; otherwise, it falls back to
the complete legal same-polarity pool.

Dense and BM25 scores are independently min--max normalized within each
selected polarity pool. Let \(\widehat d(x,u)\) and
\(\widehat b(x,u)\) denote the normalized dense and BM25 scores of
candidate \(x\) for retrieval context \(u=u_t^{(k)}\). Full MERIT ranks
candidates by
\(r(x,u)=0.75\,\widehat d(x,u)+0.25\,\widehat b(x,u)\).

The retrieval query concatenates the current question, SQL, failure
context, and error type. Dense similarity captures semantic relatedness,
whereas BM25 preserves exact overlap in question terms, schema
identifiers, failure context, and SQL changes. Positive and negative
entries are ranked separately, and MERIT retrieves at most three
positive and one negative entry. Encoder, indexing, serialization,
normalization, BM25, and tie-breaking details are provided in
Appendix~\ref{app:retrieval-details}.

\paragraph{Repair generation and validation.}
The repair prompt contains the question, schema, optional dataset
evidence, current SQL, attempt status, DBMS error, predicted failure
type, local history, and retrieved memories. Positive entries are
presented as confirmed successful directions, whereas negative entries
are presented as observed unsuccessful directions. The reference SQL,
reference rows, and corrected query are never included.

MERIT provides no separate system prompt: each template is sent as one
user message through the model's chat template. MERIT Full does not
generate local reflections. The frozen model generates one revision
\(y_t^{(k)}=P(p_t^{(k)})\) using greedy decoding; complete prompts and
memory serialization are given in Appendix~\ref{app:prompts}.

The revision is appended to the local history and evaluated by the
benchmark oracle. If it is confirmed correct, the episode terminates.
Otherwise, MERIT reclassifies the resulting failure and repeats
retrieval and generation while budget remains.

\paragraph{Episode finalization and memory update.}
If repair succeeds at step \(k^{*}\), MERIT creates one positive entry
from the final oracle-confirmed transition
\(y_t^{(k^{*}-1)}\rightarrow y_t^{(k^{*})}\). Intermediate unsuccessful
attempts remain local. If no attempt is confirmed correct after \(K\)
generations, MERIT creates at most one negative entry from the final
observed direction \(y_t^{(K-1)}\rightarrow y_t^{(K)}\).

Let \(x_t\in\{x_t^{+},x_t^{-},\varnothing\}\) denote the finalized
outcome; memory is updated only upon finalization
\begin{equation}
\mathcal{M}_{<(t+1)}
=
\mathcal{M}_{<t}
\cup
\bigl(\{x_t\}\setminus\{\varnothing\}\bigr)
\label{eq:memory-update}
\end{equation}

Episode \(t\) can therefore retrieve only experience from earlier
completed queries, while its outcome becomes visible beginning at
position \(t+1\).

\subsection{Type-Reliability-Aware Retrieval Variant}
\label{subsec:type-reliability}

We additionally evaluate a controlled MERIT variant that changes only
how the predicted type affects candidate selection and ranking. It does
not use a learned or calibrated confidence score. Instead, define
\(h(\tau)=1\) for \textsc{Syntax}, \textsc{Schema Linking}, and
\textsc{Execution}, and \(h(\tau)=0\) otherwise. These predefined
high-reliability types are intended to be supported directly by
observable DBMS diagnostics; the indicator is not estimated from
validation data. For each polarity, the variant selects
\begin{equation}
\mathcal{C}^{\mathrm{rel}}_{t,p}
=
\begin{cases}
\mathcal{T}_{t,p},
&
h(\tau_t^{(k-1)})=1
\land
|\mathcal{T}_{t,p}|\geq 3,\\
\mathcal{L}_{t,p},
& \text{otherwise}.
\end{cases}
\label{eq:reliability-filter}
\end{equation}

\begin{figure*}[t]
\centering
\includegraphics[width=\linewidth]{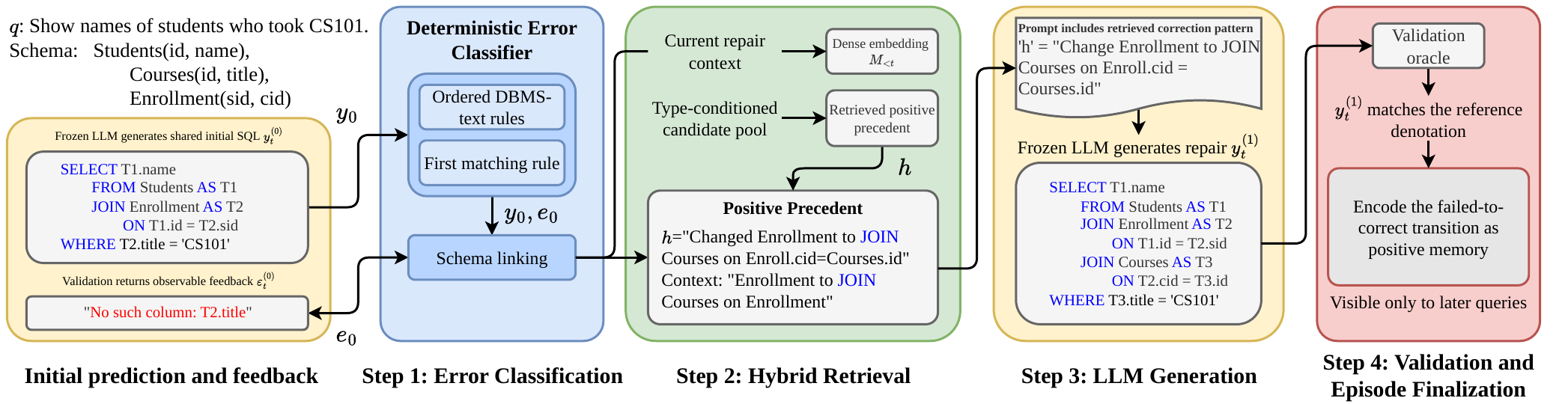}
\caption{
Example MERIT repair episode. The shared initial prediction references a
nonexistent column, producing a schema-linking failure. MERIT retrieves
an earlier correction involving the missing \texttt{Courses} join and
uses it to generate a revision. After the benchmark correctness oracle
confirms the repaired query, the failed-to-correct transition is stored
as positive memory and becomes available only to later queries.
}
\label{fig:case_study}
\end{figure*}

Lower-reliability types use the full legal pool and receive a soft
type-match bonus. Let
\(r_{\mathrm{base}}(x,u)=0.75\,\widehat r_{\mathrm{dense}}(x,u)
+0.25\,\widehat r_{\mathrm{BM25}}(x,u)\).
The variant ranks candidates by
\(r_{\mathrm{rel}}(x,u)=r_{\mathrm{base}}(x,u)
+0.10\bigl(1-h(\tau)\bigr)\mathbf{1}[\tau_x=\tau]\),
where \(u=u_t^{(k)}\) and \(\tau=\tau_t^{(k-1)}\).
The bonus is therefore applied only to lower-reliability types and is
disabled for high-reliability types, including small-pool fallback
cases. Figure~\ref{fig:case_study} illustrates a schema-linking episode. The
initial query references a nonexistent column in
\texttt{Enrollment}; MERIT retrieves an earlier correction involving
the missing \texttt{Courses} join and uses it to generate a revision.
The transition is stored as positive memory only after the benchmark
oracle confirms the repaired query.

\section{Experiments}
\label{sec:experiments}

\subsection{Experimental Setup}
\label{subsec:experimental_setup}

\paragraph{Datasets and evaluation.}
We evaluate on the development sets of Spider and BIRD, comprising
\(1{,}034\) and \(1{,}534\) queries, respectively. Spider tests
cross-domain generalization to unseen schemas, whereas BIRD contains
larger databases, more domain-specific questions, and auxiliary evidence
provided with each example. We score final predictions with the official
execution evaluators. A query is counted as correct only when its
denotation matches the evaluation target; successful execution alone is
insufficient. All prompt templates and memory-serialization formats are provided in
Appendix~\ref{app:prompts}.

\paragraph{Controlled initialization.}
All methods use the same frozen Qwen2.5-7B-Instruct backbone and the same
cached initial SQL prediction for each query. These predictions are
generated once with deterministic decoding and achieve \(58.90\%\)
execution accuracy on Spider (\(609/1{,}034\)) and \(38.40\%\) on BIRD
(\(589/1{,}534\)). Initially correct queries are finalized immediately.
For every initially incorrect query, all methods receive the same
execution feedback, decoding policy, and repair budget of \(K=7\)
generations. The comparison therefore isolates differences in repair strategy.

\paragraph{Causal stream protocol.}
We run each method under three query orders, denoted by seeds \(0\), \(1\),
and \(2\). The shared initial predictions remain fixed; only the order in
which episodes are processed changes. Within each seed, all methods
observe the same stream. Memory-based methods start from an empty store
and may retrieve only entries produced by completed episodes at earlier
stream positions. Attempts from the current query remain in its local
history and cannot enter global retrieval until the episode terminates.
Each stream order therefore induces a distinct sequence of causally
available experience.

\paragraph{Comparison methods.}
Our principal baselines isolate the contribution of cross-query memory
and its organization. \textsc{Iterative} repairs using execution feedback
and the current episode history, but maintains no global memory.
\textsc{Dynamic RAG} accumulates earlier repair experiences using the same causal eligibility rule, serialization format, and hybrid dense–BM25 ranking as MERIT, but retrieves from a single unpartitioned pool without positive/negative polarity separation or failure-type conditioning (full configuration and prompt templates are given in Appendix~\ref{sec:config_DynRAG}).
\textsc{Reflexion-style} augments repair with verbal reflections derived
from failed attempts. MERIT combines causal cross-query memory,
separate positive and negative pools, error-type conditioning, and
hybrid lexical-dense ranking. \textsc{MERIT (Type-reliability-aware)} is a controlled MERIT variant
that applies hard filtering only to predefined high-reliability failure
types and otherwise uses type agreement as a soft ranking bonus.

\paragraph{Metrics and statistical analysis.}
We report final execution accuracy and the number of initially incorrect
queries that are eventually repaired. Repair behavior is characterized
by mean repair steps per initial failure and trajectory oscillation,
while inference cost is measured by total prompt and output tokens and
LLM calls. Unless otherwise stated, results are the mean and sample
standard deviation over the three stream orders.

Because all methods are aligned on the same queries and initial
predictions, we also perform paired query-level comparisons. For the
three-seed analysis, we use a query-cluster bootstrap that resamples each
query once while keeping its outcomes across all three stream orders
together.

\subsection{Results}
\label{subsec:results}

We analyze four questions: whether cross-query memory improves stateless
repair, whether MERIT outperforms alternative memory organizations, how
robust type conditioning is, and how repair gains trade off against
inference cost.

\begin{table}[t]
\caption{
Execution accuracy over three stream orders.
Values are mean \(\pm\) sample standard deviation in percentage points.
\textbf{Bold} and \underline{underlined} denote the best and second-best
results within each dataset; tied best results are both bold.
}
\label{tab:main}
\centering
\small
\begin{tabularx}{\columnwidth}{l>{\raggedleft\arraybackslash}X}
\toprule
Method & Exec.\ accuracy \\
\midrule
\multicolumn{2}{c}{\textit{Spider}} \\
\midrule
Iterative
& \(66.34 \pm 0.14\) \\
Reflexion-style
& \(\underline{68.38 \pm 0.12}\) \\
Dyn.\ RAG
& \(\bm{69.79 \pm 0.28}\) \\
\textbf{\method}
& \(\bm{69.79 \pm 0.59}\) \\
\midrule
\multicolumn{2}{c}{\textit{BIRD}} \\
\midrule
Iterative
& \(47.35 \pm 0.04\) \\
Reflexion-style
& \(\bm{51.24 \pm 0.15}\) \\
Dyn.\ RAG
& \(48.15 \pm 0.19\) \\
\textbf{\method}
& \(\underline{48.44 \pm 0.46}\) \\
\bottomrule
\end{tabularx}
\end{table}

\begin{table}[t]
\caption{
Paired query-level comparison of MERIT against the principal baselines
and the type-reliability-aware MERIT variant.
\(\Delta\) is \method{} minus the comparison method.
Intervals are obtained by resampling queries while preserving their
outcomes across the three stream orders.
}
\label{tab:paired}
\centering
\small
\resizebox{\columnwidth}{!}{%
\begin{tabular}{llrr}
\toprule
Dataset & Comparison & \(\Delta\) pp & 95\% CI \\
\midrule
\multirow{4}{*}{Spider}
& Iterative
& \(+3.45\)
& \([+1.97,+4.96]\) \\
& Reflexion-style
& \(+1.42\)
& \([-0.19,+3.03]\) \\
& Dyn.\ RAG
& \(0.00\)
& \([-1.00,+0.97]\) \\
& MERIT (type-rel.)
& \(+0.39\)
& \([-0.23,+1.03]\) \\
\midrule
\multirow{4}{*}{BIRD}
& Iterative
& \(+1.09\)
& \([0.00,+2.17]\) \\
& Reflexion-style
& \(-2.80\)
& \([-4.15,-1.46]\) \\
& Dyn.\ RAG
& \(+0.28\)
& \([-0.43,+1.00]\) \\
& MERIT (type-rel.)
& \(-0.20\)
& \([-0.67,+0.28]\) \\
\bottomrule
\end{tabular}%
}
\end{table}

\begin{table}[t]
\caption{
Mean repair behavior and inference cost over three stream orders.
``Rep.'' is the number of initial failures repaired;
``Steps'' is repair steps per initial failure;
``Osc.'' is the oscillation rate; and ``Tok.'' is total prompt and
output tokens in millions.
\textbf{Bold} and \underline{underlined} denote the best and second-best
values within each dataset and metric.
}
\label{tab:repair-detail}
\centering
\scriptsize
\resizebox{\columnwidth}{!}{%
\begin{tabular}{lrrrrr}
\toprule
Method
& Rep.\(\uparrow\)
& Steps\(\downarrow\)
& Osc.\(\downarrow\)
& Tok.\(\downarrow\)
& Calls\(\downarrow\) \\
\midrule
\multicolumn{6}{c}{\textit{Spider}} \\
\midrule
Iterative
& 77.0
& 6.082
& 35.69
& \textbf{2.529}
& 3619 \\

Reflexion-style
& \underline{98.0}
& 5.821
& 32.69
& 6.597
& 6309 \\

Dyn.\ RAG
& \textbf{112.7}
& \underline{5.631}
& \underline{32.20}
& 5.275
& \underline{3427} \\

\textbf{\method}
& \textbf{112.7}
& \textbf{5.611}
& \textbf{31.95}
& \underline{5.274}
& \textbf{3419} \\

\midrule
\multicolumn{6}{c}{\textit{BIRD}} \\
\midrule
Iterative
& 137.3
& 6.238
& 54.52
& \textbf{9.325}
& 7429 \\

Reflexion-style
& \textbf{197.0}
& \textbf{5.971}
& \textbf{51.76}
& 22.407
& 13568 \\

Dyn.\ RAG
& 149.7
& 6.172
& 53.94
& 18.012
& \underline{7366} \\

\textbf{\method}
& \underline{154.0}
& \underline{6.128}
& \underline{53.13}
& \underline{17.569}
& \textbf{7325} \\

\bottomrule
\end{tabular}%
}
\end{table}

\paragraph{Does cross-query memory improve repair?}
On Spider, all memory-based methods improve over stateless
\textsc{Iterative} repair. MERIT and \textsc{Dynamic RAG} tie at
\(69.79\%\), followed by \textsc{Reflexion-style} at \(68.38\%\) and
\textsc{Iterative} at \(66.34\%\). MERIT's \(3.45\)-point gain over
\textsc{Iterative} has a paired \(95\%\) confidence interval of
\([1.97,4.96]\), providing clear evidence that earlier repair experience
benefits later queries. Its differences from \textsc{Dynamic RAG} and
\textsc{Reflexion-style} are not reliably resolved because both
intervals include zero. On BIRD, \textsc{Reflexion-style} performs best
at \(51.24\%\), followed by MERIT at \(48.44\%\),
\textsc{Dynamic RAG} at \(48.15\%\), and \textsc{Iterative} at
\(47.35\%\). MERIT's \(1.09\)-point gain over \textsc{Iterative} has an
interval of \([0.00,2.17]\), indicating weaker evidence than on Spider.

\paragraph{Does structured memory outperform alternative memories?}
MERIT and \textsc{Dynamic RAG} are not reliably separated: they tie on
Spider, and MERIT leads by only \(0.28\) points on BIRD. Thus, the
results do not establish an aggregate benefit from MERIT's polarity
separation and hard type conditioning over untyped dynamic retrieval.
MERIT exceeds \textsc{Reflexion-style} by \(1.42\) points on Spider, but
the paired interval \([-0.19,3.03]\) includes zero. On BIRD,
\textsc{Reflexion-style} leads MERIT by \(2.80\) points, with a
MERIT-minus-Reflexion interval of \([-4.15,-1.46]\), making it the
strongest BIRD method.

\paragraph{How robust is error-type conditioning?}
Hard and type-reliability-aware filtering are also not reliably
distinguishable. Hard filtering changes mean accuracy by \(+0.39\)
points on Spider and \(-0.20\) points on BIRD, with intervals of
\([-0.23,1.03]\) and \([-0.67,0.28]\), respectively. The opposite
directions and zero-crossing intervals indicate that neither policy is
uniformly better, supporting the use of error type as a coarse retrieval prior.

\paragraph{How does memory affect repair behavior?}
Memory primarily increases the number of initially failed queries that
are eventually repaired. On Spider, MERIT and \textsc{Dynamic RAG} each
repair \(112.7\) failures on average, compared with \(98.0\) for
\textsc{Reflexion-style} and \(77.0\) for \textsc{Iterative}. MERIT also
has the shortest trajectories and lowest oscillation, although its
differences from \textsc{Dynamic RAG} are small. On BIRD,
\textsc{Reflexion-style} repairs the most failures (\(197.0\)) and
produces the shortest, least oscillatory trajectories. MERIT ranks
second with \(154.0\) repaired failures, compared with \(149.7\) for
\textsc{Dynamic RAG} and \(137.3\) for \textsc{Iterative}. Across both
datasets, differences in trajectory length and oscillation are smaller
than differences in repair success.

\paragraph{What is the computational trade-off?}
Higher repair success comes with greater token use. On Spider, MERIT
uses \(5.274\) million tokens and \(3{,}419\) calls, compared with
\(2.529\) million tokens and \(3{,}619\) calls for
\textsc{Iterative}. \textsc{Dynamic RAG} has nearly identical cost,
whereas \textsc{Reflexion-style} requires \(6.597\) million tokens and
\(6{,}309\) calls. On BIRD, MERIT uses \(17.569\) million tokens and
\(7{,}325\) calls, compared with \(9.325\) million tokens and
\(7{,}429\) calls for \textsc{Iterative}. \textsc{Reflexion-style}
achieves the highest accuracy but is also the most expensive, requiring
\(22.407\) million tokens and \(13{,}568\) calls. Thus, MERIT is less
expensive than \textsc{Reflexion-style}, but does not reduce absolute
token consumption relative to stateless repair.
\section{Ablation Analysis}
\label{sec:ablation}

We ablate memory polarity, error-type conditioning, lexical-dense
ranking, and schema-local retrieval. All variants use the same frozen
backbone, cached initial predictions, three causal stream orders, and
repair budget \(K=7\). Table~\ref{tab:ablation} reports mean execution
accuracy and sample standard deviation.

\begin{table}[t]
\caption{
Execution accuracy for MERIT ablations over three stream orders.
\(\Delta\) is the change in percentage points relative to full MERIT.
\textbf{Bold} and \underline{underlined} denote the best and second-best
results within each dataset; tied best results are both bold.
}
\label{tab:ablation}
\centering
\small
\resizebox{\columnwidth}{!}{%
\begin{tabular}{lcc|cc}
\toprule
\multirow{2}{*}{Variant}
& \multicolumn{2}{c|}{Spider}
& \multicolumn{2}{c}{BIRD} \\
\cmidrule(lr){2-3}\cmidrule(lr){4-5}
& Exec. & \(\Delta\)
& Exec. & \(\Delta\) \\
\midrule

\textbf{\method{} Full}
& \underline{\(69.79 \pm 0.59\)} & --
& \underline{\(48.44 \pm 0.46\)} & -- \\

Positive only
& \(69.66 \pm 0.53\) & \(-0.13\)
& \(48.35 \pm 0.72\) & \(-0.09\) \\

No type filter
& \(69.21 \pm 0.15\) & \(-0.58\)
& \textbf{\(48.57 \pm 0.49\)} & \(+0.13\) \\

No dense rerank
& \(68.67 \pm 0.29\) & \(-1.12\)
& \textbf{\(48.57 \pm 0.46\)} & \(+0.13\) \\

No BM25
& \textbf{\(70.37 \pm 0.20\)} & \(+0.58\)
& \(48.04 \pm 0.07\) & \(-0.40\) \\

Random same type
& \(69.70 \pm 0.44\) & \(-0.09\)
& \(48.39 \pm 0.73\) & \(-0.05\) \\

Cross-database only
& \(68.86 \pm 0.29\) & \(-0.93\)
& \(45.70 \pm 0.39\) & \(-2.74\) \\

\bottomrule
\end{tabular}%
}
\end{table}

\paragraph{How much does polarity contribute?}
Removing negative memory changes accuracy only modestly:
\(-0.13\) points on Spider and \(-0.09\) points on BIRD. These margins
are small relative to variation across stream orders, indicating that
verified corrections provide most of the useful signal. Negative entries
still offer a limited warning against previously unsuccessful repair
directions, but the results do not identify polarity separation as the
principal source of MERIT's gains. Negative memory is expected to help most when an unsuccessful direction recurs across similar failures - for instance, a plausible but schema-invalid join pattern that the model might otherwise retry. Because the positive pool already narrows retrieval to a verified repair direction, negative entries mainly act as a redundant filter, which is consistent with their modest measured contribution.

\paragraph{When does error typing help?}
Type-conditioned retrieval is more useful on Spider. Removing the type
filter lowers accuracy by \(0.58\) points, while Random Same Type remains
only \(0.09\) points below full MERIT. Thus, narrowing retrieval to a
plausible failure class already provides substantial structure, even
before fine-grained ranking. Dense relevance remains important within
that class: removing dense reranking produces the largest Spider decline
among the retrieval ablations, \(1.12\) points.

BIRD presents a different regime. No Type Filter and No Dense Rerank
both reach \(48.57\%\), only \(0.13\) points above full MERIT, while
Random Same Type is \(0.05\) points below it. These differences are
small relative to their standard deviations and do not establish a
preferred retrieval policy. The type-reliability-aware comparison in
Table~\ref{tab:paired} leads to the same conclusion: hard filtering is
favored by \(0.39\) points on Spider and disfavored by \(0.20\) points
on BIRD, but both confidence intervals include zero. Error type is
therefore useful as an organizing prior, particularly on Spider, rather
than as a universally reliable retrieval boundary.

\paragraph{How do lexical and dense ranking interact?}
The two retrieval channels behave differently across benchmarks. On
Spider, removing dense reranking lowers accuracy by \(1.12\) points,
whereas removing BM25 raises the mean by \(0.58\) points and reduces
variance. This pattern suggests that semantic similarity is better
aligned with transferable repair structure, while exact lexical overlap
can favor surface-level matches. However, the paired interval for the
No-BM25 comparison includes zero, so the evidence does not establish
that lexical retrieval is reliably harmful.

On BIRD, removing BM25 lowers accuracy by \(0.40\) points, while
removing dense reranking changes the mean by only \(+0.13\) points.
Exact overlap in table names, columns, and domain terminology may thus
be more informative for BIRD's larger and more specialized schemas.
A fixed lexical-dense mixture provides a common retrieval policy, but
the ablations show that the relative value of its channels is
environment dependent.

\paragraph{How transferable is repair experience across schemas?}
Schema locality produces the clearest and most consistent effect.
Restricting retrieval to memories from other databases lowers accuracy
by \(0.93\) points on Spider and \(2.74\) points on BIRD. Repair
patterns do transfer across schemas, but cross-database analogies do not
fully replace experience from the target database. Same-database
memories can preserve recurring join paths, table relationships, and
domain terminology that are difficult to recover from question
similarity or failure type alone. The larger BIRD degradation highlights the importance of schema-local
experience in domain-specific databases.

\paragraph{Error-classifier validation.}
Table~\ref{tab:error-diagnostics} characterizes the operating regime of
the deterministic classifier over \(425\) Spider and \(945\) BIRD
initial failures. Spider is dominated by \textsc{Result Mismatch}
(\(84.71\%\)), whereas BIRD contains substantially more
\textsc{Schema Linking} and other failure types. Despite this difference,
accuracy remains near \(48\%\) on both datasets, with Macro-F1 scores of
\(0.317\) and \(0.355\). These results show that the labels provide useful
but imperfect routing information. We therefore use error type as a coarse retrieval prior, retain a
same-polarity fallback when the typed pool is small, and separately
evaluate a type-reliability-aware policy that relaxes hard filtering for
less reliable categories.

\begin{table}[t]
\caption{
Initial-failure composition and provisional classifier performance.
Shares are percentages. ``Result mismatch'' denotes executable SQL with
an incorrect denotation; ``Other'' aggregates Aggregation, Execution,
Syntax, and Unknown.
}
\label{tab:error-diagnostics}
\centering
\small
\resizebox{\columnwidth}{!}{%
\begin{tabular}{@{}lccccc@{}}
\toprule
Dataset
& \shortstack{Result\\mismatch}
& \shortstack{Schema\\linking}
& Other
& \shortstack{Acc.\\(\%)}
& Macro-F1 \\
\midrule
Spider & 84.71 & 12.24 & 3.06 & 48.71 & 0.317 \\
BIRD   & 57.99 & 33.65 & 8.36 & 48.10 & 0.355 \\
\bottomrule
\end{tabular}%
}
\end{table}

\section{Conclusion}

We introduced \method{}, a training-free framework that learns from
execution-guided Text-to-SQL repair through causal cross-query memory.
\method{} clearly improves over stateless repair on Spider, while its
BIRD gain is smaller; it is not reliably separated from Dynamic RAG on
either benchmark. The ablations show that polarity contributes modestly,
retrieval design is dataset dependent, and schema-local experience is
consistently valuable. These results support memory-based improvement
without parameter updates, while showing that no single memory
organization is uniformly best.

\section{Limitations}

MERIT assumes access to a denotation-level correctness signal during
repair. This is appropriate for controlled benchmark evaluation, but it
is stronger than the feedback available in many deployed databases,
where an agent may observe only execution errors, incomplete tests, or
delayed user confirmation. Our experiments also use a single frozen
backbone, Qwen2.5-7B-Instruct, and the development sets of Spider and
BIRD. We therefore do not establish that the same gains or trade-offs
hold for other model families, multilingual or conversational
Text-to-SQL, or production databases with different execution and
security constraints.

The method is sensitive to how experience is accumulated and organized.
Memory contents depend on query order, and three stream orders capture
only a limited portion of this variation. The deterministic classifier
is also coarse: its population-weighted validation accuracy is
approximately \(48\%\), and the broad Result Mismatch category combines
multiple underlying semantic errors. This limits how strongly the
results can support hard type-conditioned retrieval and helps explain
why relaxed filtering remains competitive on BIRD. We also fix the
retrieval weights, memory allocation, and fallback policy across both
benchmarks.

Finally, MERIT is not cheaper than stateless repair in total token use.
Although it repairs more failures and makes slightly fewer model calls,
retrieved memories lengthen each prompt, resulting in substantially
higher token consumption. Our accounting includes LLM prompt and output
tokens but excludes embedding computation, indexing, retrieval latency,
and memory-maintenance overhead.

\bibliography{custom}

\appendix
\section{MERIT Correction-Accumulation Loop}
Algorithm~\ref{alg:merit} restates the causal correction-accumulation
process of Section~3 as explicit pseudocode, tracing one query from its
shared initial prediction through classification, retrieval, repair, and
memory finalization.
\label{app:algorithm}
\begin{algorithm*}[t]
\caption{MERIT causal correction-accumulation loop}
\label{alg:merit}
\begin{algorithmic}[1]
\Require Ordered stream
\(\{(q_t,S_t,a_t,D_t,y_t^{(0)})\}_{t=1}^{N}\),
parser \(P\), evaluator \(V\), budget \(K\)
\State \(\mathcal{M}^{+}\gets\varnothing\);
       \(\mathcal{M}^{-}\gets\varnothing\)
\For{\(t=1,\ldots,N\)}
    \State Evaluate shared prediction \(y_t^{(0)}\)
    \If{\(y_t^{(0)}\) is correct}
        \State Finalize without creating repair memory
        \State \textbf{continue}
    \EndIf
    \State \(H_t\gets[y_t^{(0)}]\);
           \(\mathrm{solved}\gets\textbf{false}\)
    \For{\(k=1,\ldots,K\)}
        \State Classify the current failure type \(\tau_t^{(k-1)}\)
        \State Construct the legal earlier-memory pool
        \State Retrieve up to three positive and one negative entries
        \State Build prompt from current context, \(H_t\), and retrieval
        \State Generate one repair \(y_t^{(k)}\gets P(p_t^{(k)})\)
        \State Append \(y_t^{(k)}\) to \(H_t\) and evaluate it
        \If{\(y_t^{(k)}\) is correct}
            \State Encode final failed-to-correct transition as \(x_t^{+}\)
            \State \(\mathrm{solved}\gets\textbf{true}\)
            \State \textbf{break}
        \EndIf
    \EndFor
    \If{\(\neg\mathrm{solved}\)}
        \State Encode final unsuccessful direction as \(x_t^{-}\)
    \EndIf
    \State Insert finalized \(x_t\) into memory
           \Comment{visible only for positions \(>t\)}
\EndFor
\end{algorithmic}
\end{algorithm*}

\section{Prompt Templates}
\label{app:prompts}

MERIT supplies no separate system prompt. Each template below is passed
as a single user message through the model's native chat template.
Placeholders enclosed in braces are replaced at inference time. Spider
omits all BIRD-specific evidence blocks. MERIT Full does not generate
local reflections, so the corresponding field in its repair prompt is
always rendered as \texttt{(none)}.

\subsection{Shared Initial SQL Generation}

The same initial-generation prompt is used to produce the cached SQL
prediction shared by all repair methods. The prompt-version identifier
is \texttt{spider-initial-v3} or \texttt{bird-initial-v3}, depending on
the benchmark.

\begin{lstlisting}[style=prompt,
caption={Shared initial SQL-generation prompt.},
label={lst:initial-prompt}]
PROMPT_VERSION: {spider-initial-v3 | bird-initial-v3}

You are an expert SQLite developer. Produce one SQL query for the question.

DATABASE SCHEMA:
{DATABASE_SCHEMA}

{BIRD_CONTEXT_IF_APPLICABLE}

QUESTION:
{QUESTION}

RULES:
- Use only exact table and column names from the schema.
- Do not create table or column aliases with AS.
- Return exactly one query; do not provide alternatives.
- Put the final SQL between <answer> and </answer> tags.
\end{lstlisting}

For BIRD, \texttt{BIRD\_CONTEXT\_IF\_APPLICABLE} is replaced with the
following block. Spider receives no corresponding block.

\begin{lstlisting}[style=prompt,
caption={BIRD-specific evidence block for initial generation.},
label={lst:bird-initial-block}]
EXTERNAL KNOWLEDGE / EVIDENCE:
{BIRD_EVIDENCE}

BIRD RULES:
- Implement the evidence formula or computation exactly.
- Wrap column names containing spaces or special characters in backticks, for example `Column Name`.
\end{lstlisting}

\subsection{MERIT Repair Generation}

Listing~\ref{lst:merit-repair-prompt} gives the central repair prompt
used after an unsuccessful SQL attempt. The prompt contains the current
failure signal, complete local attempt history, and retrieved positive
and negative memories. The prompt-version identifier is
\texttt{spider-repair-v3} or \texttt{bird-repair-v3}.

\begin{lstlisting}[style=prompt,
caption={MERIT repair prompt.},
label={lst:merit-repair-prompt}]
PROMPT_VERSION: {spider-repair-v3 | bird-repair-v3}

You are an expert SQLite developer repairing an unsuccessful query.

CONFIRMED SUCCESSFUL REPAIR DIRECTIONS:
{RETRIEVED_POSITIVE_MEMORIES_OR_(none)}

OBSERVED FAILED DIRECTIONS:
{RETRIEVED_NEGATIVE_MEMORIES_OR_(none)}

LOCAL REFLECTIONS FROM THIS EPISODE:
(none)

CURRENT FEEDBACK:
Status: {DENOTATION_MISMATCH | EXECUTION_ERROR | TIMEOUT}
Current error type: {ERROR_TYPE}
DB error: {DATABASE_ERROR_OR_(none)}

DATABASE SCHEMA:
{DATABASE_SCHEMA}

{BIRD_CONTEXT_IF_APPLICABLE}

QUESTION:
{QUESTION}

LOCAL ATTEMPT HISTORY:
{ALL_PREVIOUS_SQL_ATTEMPTS}

REPAIR RULES:
- Produce a new SQL query rather than repeating a prior attempt.
- Use only exact table and column names from the schema.
- Treat failed directions only as observed evidence; do not invent a reason.
- Put exactly one final SQL query between <answer> and </answer> tags.
\end{lstlisting}

For BIRD, the schema section is followed by the same evidence block
shown in Listing~\ref{lst:bird-initial-block}.

\subsection{Memory Serialization}

Retrieved positive memories are rendered as verified
failed-to-correct transitions:

\begin{lstlisting}[style=prompt,
caption={Positive-memory serialization.},
label={lst:positive-memory}]
[Confirmed successful repair {INDEX}]
Entry ID: {ENTRY_ID}
Error type: {ERROR_TYPE}
Failure context: {PREVIOUS_FAILURE_CONTEXT}
Observed successful direction: {FAILED_SQL_TO_CORRECT_SQL_TRANSITION}
SQL delta: {SQL_CHANGE}
\end{lstlisting}

Negative memories describe only an observed unsuccessful direction and
its recorded outcome:

\begin{lstlisting}[style=prompt,
caption={Negative-memory serialization.},
label={lst:negative-memory}]
[OBSERVED FAILED DIRECTION {INDEX}]
Entry ID: {ENTRY_ID}
Error type: {ERROR_TYPE}
Failure context: {FAILURE_CONTEXT}
Attempted SQL delta: {SQL_CHANGE}
Observed outcome: {OUTCOME}
Observed DB error: {DATABASE_ERROR}
\end{lstlisting}

The complete local trajectory is serialized in chronological order.
Because repair is invoked only after an unsuccessful attempt, every SQL
query included in this block has already been observed to be incorrect.

\begin{lstlisting}[style=prompt,
caption={Local attempt-history serialization.},
label={lst:attempt-history}]
[Attempt 1 - observed unsuccessful]
{INITIAL_SQL}

[Attempt 2 - observed unsuccessful]
{FIRST_REPAIR_SQL}

...

[Attempt N - observed unsuccessful]
{LATEST_REPAIR_SQL}
\end{lstlisting}

\subsection{Reflexion-Style Baseline}

The following prompt is used only by the Reflexion-style baseline and
is not used by MERIT Full. It asks the frozen model to summarize the
observed trajectory without generating another SQL query or asserting
an unobserved failure cause. The prompt-version identifier is
\texttt{spider-reflection-v3} or \texttt{bird-reflection-v3}.

\begin{lstlisting}[style=prompt,
caption={Reflection-generation prompt used by the Reflexion-style baseline.},
label={lst:reflection-prompt}]
PROMPT_VERSION: {spider-reflection-v3 | bird-reflection-v3}

Write a concise debugging reflection using only the observed attempt outcomes.
Do not claim an unobserved cause and do not produce the next SQL query.

DATABASE SCHEMA:
{DATABASE_SCHEMA}

{BIRD_CONTEXT_IF_APPLICABLE}

QUESTION:
{QUESTION}

CURRENT ERROR TYPE: {ERROR_TYPE}

ATTEMPTS AND OBSERVED OUTCOMES:
[Attempt 1]
SQL: {SQL_1}
Outcome: status={STATUS_1}; db_error={ERROR_1_OR_(none)}

[Attempt 2]
SQL: {SQL_2}
Outcome: status={STATUS_2}; db_error={ERROR_2_OR_(none)}

...

Put the reflection between <reflection> and </reflection> tags.
\end{lstlisting}

\section{Deterministic Online Error Classifier}
\label{app:classifier-rules}

The online classifier is conservative and uses only observed DBMS
diagnostics and execution status. It does not inspect the SQL abstract
syntax tree or infer semantic errors from query structure. Rules are
evaluated in the order shown in Table~\ref{tab:classifier-rules}, and
the first matching rule determines the output.

\begin{table*}[t]
\caption{Ordered deterministic rules used by the online failure
classifier.}
\label{tab:classifier-rules}
\centering
\small
\begin{tabular}{@{}p{0.42\textwidth}p{0.22\textwidth}p{0.25\textwidth}@{}}
\toprule
Observed DBMS status or text & Type & Subtype \\
\midrule
\texttt{no such table:}
& Schema Linking & Missing Table \\

\texttt{no such column:}
& Schema Linking & Missing Column \\

\texttt{no such view:}
& Schema Linking & Missing View \\

\texttt{ambiguous column name:}
& Schema Linking & Ambiguous Column \\

\texttt{no such function:}
& Syntax & Unknown Function \\

\texttt{syntax error}, \texttt{incomplete input}, or
\texttt{unrecognized token}
& Syntax & Parse Failure \\

\texttt{misuse of aggregate} or
\texttt{aggregate functions are not allowed}
& Aggregation & DBMS Aggregate Misuse \\

\texttt{datatype mismatch} or \texttt{type mismatch}
& Filter/Value & Type Mismatch \\

Attempted write under read-only execution
& Execution & Read Only Violation \\

Database not found
& Execution & DB Not Found \\

\textsc{Timeout} status or text containing \texttt{timed out}
& Execution & Timeout \\

Other nonempty DBMS error
& Unknown & Unknown \\

Executable query with incorrect denotation
& Result Mismatch & Unknown \\
\bottomrule
\end{tabular}
\end{table*}

The classifier does not diagnose join, ordering, limit, aggregation
logic, or other semantic subtypes from SQL shape. Any broader semantic
taxonomy used for manual analysis is therefore distinct from the labels
produced by the online classifier.

\section{Retrieval and Implementation Details}
\label{app:retrieval-details}

Table~\ref{tab:retrieval-config} reports the executed configuration used
for the main MERIT experiments and its retrieval-policy variants.

\begin{table*}[t]
\caption{Executed retrieval, memory, generation, and feedback settings.}
\label{tab:retrieval-config}
\centering
\small
\begin{tabular}{@{}p{0.22\textwidth}p{0.72\textwidth}@{}}
\toprule
Component & Setting \\
\midrule
Dense encoder
& \texttt{BAAI/bge-large-en-v1.5}\\

Encoding
& Sentence Transformers 3.2.1 with \(L_2\)-normalized embeddings \\

Dense similarity
& Inner product between normalized vectors, equivalent to cosine
similarity \\

Dense index
& FAISS \texttt{IndexFlatIP}, FAISS GPU 1.9.0.0 \\

BM25 implementation
& Custom in-project implementation; the pinned \texttt{rank-bm25}
package is not called \\

BM25 tokenization
& Lowercasing followed by the regular expression
\texttt{[a-z0-9\_]+} \\

BM25 parameters
& \(k_1=1.5\) and \(b=0.75\) \\

Score normalization
& Independent min-max normalization within each selected candidate
polarity pool \\

Main ranking
& \(0.75\widehat{s}_{\mathrm{dense}}
 +0.25\widehat{s}_{\mathrm{BM25}}\) \\

Retrieval limits
& At most three positive and one negative memory \\

Typed threshold
& Three same-type entries, applied independently within each polarity \\

Prompt format
& \texttt{merit-prompts-v3} \\

Decoding
& Greedy decoding, sampling disabled, and temperature \(0\) \\

Repair budget
& Seven repair generations after the shared initial prediction \\
\bottomrule
\end{tabular}
\end{table*}

\paragraph{Dense retrieval.}
The dense encoder is
\texttt{BAAI/bge-large-en-v1.5} at the revision listed in
Table~\ref{tab:retrieval-config}. Sentence Transformers 3.2.1 produces
\(L_2\)-normalized vectors, which are stored in a FAISS
\texttt{IndexFlatIP} index. Dense relevance is the inner product of the
normalized query and memory vectors and is therefore equivalent to
cosine similarity.

The dense retrieval query concatenates the current question, current
SQL, current failure context, and predicted error type. Dense memory
text concatenates the source question, source schema, failure context,
normalized SQL transformation, error type, observed outcome, and
observed DBMS error.

\paragraph{Lexical retrieval.}
The lexical channel uses a custom BM25 implementation. Both retrieval
queries and memory documents are lowercased and tokenized with
\texttt{[a-z0-9\_]+}. For query \(q\) and candidate memory \(m\), the lexical score is

\begin{equation}
\operatorname{BM25}(q,m)
=
\sum_{w\in q}
\operatorname{IDF}(w)\,g(w,m)
\label{eq:bm25}
\end{equation}
where the term-frequency saturation factor is

\begin{equation}
g(w,m)
=
\frac{\operatorname{tf}(w,m)(k_1+1)}
{\operatorname{tf}(w,m)
+k_1\left(1-b+b|m|/\bar{\ell}\right)}
\label{eq:bm25-tf}
\end{equation}
and the inverse-document-frequency term is

\begin{equation}
\operatorname{IDF}(w)
=
\log\left(
1+
\frac{N-\operatorname{df}(w)+0.5}
{\operatorname{df}(w)+0.5}
\right)
\label{eq:bm25-idf}
\end{equation}
where \(N\) is the number of memories in the selected candidate
polarity pool, \(\operatorname{df}(w)\) is the number of memories
containing token \(w\), \(\operatorname{tf}(w,m)\) is its frequency in
\(m\), and \(\overline{\ell}\) is the mean document length in that pool.
We use \(k_1=1.5\) and \(b=0.75\). Corpus statistics are recomputed after
candidate selection, independently for the positive and negative pools.

BM25 memory text contains the source question, failure context, SQL
delta, error type, observed outcome, observed DBMS error, and observed
successful direction when one exists.

\paragraph{Normalization and hybrid ranking.}
Dense and BM25 scores are normalized independently within each selected
candidate polarity pool. For score function \(s\), candidate \(m\), and
pool \(\mathcal{C}_{t,p}\), normalization is
\begin{equation}
\widehat{s}(m)
=
\begin{cases}
\dfrac{s(m)-s_{\min}}
{s_{\max}-s_{\min}},
& s_{\max}>s_{\min},\\[6pt]
0,
& \text{otherwise},
\end{cases}
\label{eq:minmax-normalization}
\end{equation}
where \(s_{\min}\) and \(s_{\max}\) are computed over
\(\mathcal{C}_{t,p}\). Full MERIT then ranks candidates using
\(0.75\widehat{s}_{\mathrm{dense}}
+0.25\widehat{s}_{\mathrm{BM25}}\), as defined in the full-MERIT ranking function. Positive and negative memories are
ranked separately, and retrieval returns at most three positive and one
negative entry.

For the type-reliability-aware variant, the \(0.10\) type-match bonus is
applied only to lower-reliability predicted types, as specified by the type-reliability-aware ranking function. Ranking ties are resolved by
descending dense score, descending BM25 score, descending type-match
score, and finally ascending deterministic memory ID.

\paragraph{Memory and generation configuration.}
Memory entries use format version 3 and are stored in separate positive
and negative JSONL files. Positive entries contain an oracle-confirmed
failed-to-correct transition, whereas negative entries contain the final
observed unsuccessful direction from an unresolved episode. Entries are
indexed only after episode finalization.

All methods use the \texttt{merit-prompts-v3} prompt family and the same
greedy decoding configuration. Sampling is disabled, temperature is
zero, and each initially incorrect query receives at most seven repair
generations after the shared initial prediction. The complete prompt and
memory-serialization templates are provided in
Appendix~\ref{app:prompts}.

\section{Configuration and Prompt Templates for \textsc{Dynamic RAG}}
\label{sec:config_DynRAG}

\textsc{Dynamic RAG} reuses Full MERIT's causal eligibility rule, memory
serialization format, dense encoder, BM25 implementation, hybrid ranking
weights, decoding configuration, and repair budget without modification
(Appendix~\ref{app:retrieval-details}). Table~\ref{tab:dynrag-config}
reports only the settings that differ from Full MERIT; all other executed
values follow Table~\ref{tab:retrieval-config}.

\begin{table*}[t]
\caption{Retrieval and memory settings for \textsc{Dynamic RAG}, reported
only where they differ from Full MERIT.}
\label{tab:dynrag-config}
\centering
\small
\begin{tabular}{@{}p{0.22\textwidth}p{0.72\textwidth}@{}}
\toprule
Component & Setting \\
\midrule
Memory pool structure
& A single causal legal pool per repair step; candidates are not
partitioned into separate positive and negative pools before ranking \\

Failure-type conditioning
& Disabled. The deterministic classifier (Appendix~\ref{app:classifier-rules})
still labels the current failure for display in the repair prompt, but
this label does not filter or rerank candidates \\

Ranking
& \(0.75\,\widehat{s}_{\mathrm{dense}} + 0.25\,\widehat{s}_{\mathrm{BM25}}\),
identical to the full-MERIT ranking function with the type-match term
fixed at zero \\

Retrieval limit
& At most \texttt{max\_positive + max\_negative = 4} entries per repair
step, selected jointly from the combined pool \\

Negative-memory creation
& Disabled at episode finalization; only oracle-confirmed
failed-to-correct transitions are written to memory \\

Prompt templates
& Identical to the shared initial-generation prompt
(Listing~\ref{lst:initial-prompt}) and the MERIT repair prompt
(Listing~\ref{lst:merit-repair-prompt}); no separate prompt version is
defined for this baseline \\

Memory serialization
& Identical to Listings~\ref{lst:positive-memory}-\ref{lst:attempt-history};
the negative-memory block is never populated for this baseline \\

Causal eligibility, encoder, BM25 parameters, decoding, repair budget
& Identical to Full MERIT (Table~\ref{tab:retrieval-config}) \\
\bottomrule
\end{tabular}
\end{table*}

\paragraph{Memory pool and polarity.}
Candidate construction applies the same four causal checks as Full MERIT
(Appendix~\ref{app:retrieval-details}): an entry is legal only if its
source episode is finalized, its source stream position precedes the
current position, and it was not produced by the current query. Unlike
Full MERIT, this filtering is not further split by polarity, so the
retrieval mechanism itself does not distinguish confirmed corrections from
observed failed directions. In practice, however, \textsc{Dynamic RAG}'s
finalization step never creates a negative entry, so the legal pool is
populated only with oracle-confirmed positive transitions. \textsc{Dynamic
RAG} therefore differs from Full MERIT in both respects described in the
main text: it neither partitions memory by polarity nor conditions
retrieval on failure type, and every retrieved entry is in practice a
verified correction.

\paragraph{Ranking.}
Within the combined pool, dense and BM25 scores are computed and
min-max normalized exactly as in Full MERIT
(Appendix~\ref{app:retrieval-details}), and the type-match bonus is fixed
at zero because \textsc{Dynamic RAG} is untyped. The four highest-ranked
entries are returned, using the same tie-break order as Full MERIT
(descending dense score, then descending BM25 score, then ascending
deterministic memory ID).

\paragraph{Prompt templates.}
\textsc{Dynamic RAG} issues exactly the shared initial-generation prompt
(Listing~\ref{lst:initial-prompt}) and the MERIT repair prompt
(Listing~\ref{lst:merit-repair-prompt}) under the same prompt-version
identifiers, \texttt{spider-repair-v3} or \texttt{bird-repair-v3}; the
framework defines no separate template family for this baseline. Because
its memory contains no negative entries, the
\texttt{OBSERVED FAILED DIRECTIONS} field of every repair prompt renders
as \texttt{(none)}, while \texttt{CONFIRMED SUCCESSFUL REPAIR DIRECTIONS}
is filled with up to four retrieved transitions using the same
positive-memory serialization template as Full MERIT
(Listing~\ref{lst:positive-memory}). The
\texttt{LOCAL REFLECTIONS FROM THIS EPISODE} field remains
\texttt{(none)}, and the current-feedback, schema, question, and
local-attempt-history sections are constructed identically to Full MERIT.

\paragraph{Relation to Full MERIT.}
\textsc{Dynamic RAG} isolates the contribution of MERIT's structured
memory organization: it shares MERIT's causal cross-query availability,
serialization, and hybrid ranking weights, but removes both
positive/negative polarity separation and error-type conditioning by
retrieving from one unpartitioned, untyped pool. This configuration
directly supports the comparison in Table~\ref{tab:classifier-rules}
alternative and Section~4.2, where MERIT and \textsc{Dynamic RAG} are
found not to be reliably separated on either benchmark.

\end{document}